%% file: opt-tar-neurips_2023.tex
\documentclass{article}

\usepackage[preprint]{neurips_2023}

\input{packages}

\title{Comparing Optimization Targets for Contrast-Consistent Search}

\author{%
  Hugo Fry \\
  Independent \\
  hugo.fry@fryfamily.co.uk
  \And
  Seamus Fallows \\
  Independent \\
  seamusfallows1@gmail.com
  \And
  Ian Fan \\
  Independent
  \And
  Jamie Wright \\
  Oxford University
  \And
  Nandi Schoots \\
  King's College London %and \\Imperial College London
}

\begin{document}

\maketitle

\input{write-up-sections/Optimization-Target}

% \newpage

% Need to comment this out for anonymous submission!
% \section*{Acknowledgements}

% AI Safety Hub

\bibliographystyle{plainnat}
\bibliography{references}

\newpage
\input{write-up-sections/appendix-opt}

%%%%%%%%%%%%%%%%%%%%%%%%%%%%%%%%%%%%%%%%%%%%%%%%%%%%%%%%%%%%

\end{document}

%% file: packages.tex
\usepackage[utf8]{inputenc} % allow utf-8 input
\usepackage[T1]{fontenc}    % use 8-bit T1 fonts
\usepackage{hyperref}       % hyperlinks
\usepackage{url}            % simple URL typesetting
\usepackage{booktabs}       % professional-quality tables
\usepackage{amsfonts}       % blackboard math symbols
\usepackage{nicefrac}       % compact symbols for 1/2, etc.
\usepackage{microtype}      % microtypography
\usepackage{xcolor}         % colors
\usepackage{colortbl}       % colours for table in experiment AC
\usepackage{amsmath}
\usepackage{graphicx}
\usepackage{multirow}

\usepackage{tikz}
\usepackage{subcaption}

\usepackage[backgroundcolor=white, textsize=small, textwidth=20mm]{todonotes}

\usepackage{amsthm}
\usepackage{amsmath}

%% file: write-up-sections/Optimization-Target.tex
\begin{abstract}
We investigate the optimization target of Contrast-Consistent Search (CCS), which aims to recover the internal representations of truth of a large language model. We present a new loss function that we call the Midpoint-Displacement (MD) loss function. We demonstrate that for a certain hyper-parameter value this MD loss function leads to a prober with very similar weights to CCS. We further show that this hyper-parameter is not optimal and that with a better hyper-parameter the MD loss function attains a higher test accuracy than CCS.

\end{abstract}

\section{Introduction}

The increased deployment of large language models to real-world applications has continued to highlight the discrepancy between the growing capabilities of language models and our limited understanding of their behavior~\citep{openai2023gpt4}.

It is widely believed that language models have internal representations encoding knowledge of the world~\citep{DBLP:conf/iclr/0002HBVPW23}. 
In order to deploy such systems safely, a better understanding of these internal representations is needed.
In particular, language models will often confidently output false statements or `hallucinate’ in a number of settings~\citep{LLM-hallucination}. It is therefore important to develop techniques aimed at discovering internal representations of truth.

Mechanistic interpretability aims to reverse engineer the algorithms that the model weights implement at the level of individual neurons~\citep{cammarata2020,wangnd}. 
Other interpretability work more broadly aims to develop automatic techniques for probing and modifying human-interpretable concepts from a model’s activations~\citep{nand,wangnd}.

Recent work in the field of interpretability has presented contrast-consistent search (CCS)~\citep{ogpaper} as an unsupervised method for extracting knowledge from the hidden states of large language models. It is able to learn the truth value of statements using the negation consistency property of truth: a statement and its negation must have opposite truth values. By comparing the model representations corresponding to these two statements, one can find a direction in activation space that satisfies this consistency constraint.

In this paper, we investigate the optimization target of CCS. Our contributions are as follows:
\begin{itemize}
    \item In Section \ref{sec:clarifying-CCS}, we present some conceptual clarifications on how CCS classifies truth and how it should be interpreted.
    \item In Section \ref{sec:MD}, we present a heuristic explanation of CCS's optimization target. Motivated by this explanation, we introduce the Midpoint-Displacement (MD) loss function as a proxy optimisation target for CCS. 
    \item In Section \ref{sec:comparison-MD-CCS}, we present an experimental comparison of many loss functions across a number of datasets and models. In particular, we demonstrate that, for a certain choice of the hyper-parameter, the truth direction found by the MD loss function is a good proxy for CCS, as measured by cosine similarity. 
    \item Further, we tentatively show that this hyper-parameter is not optimal and that with a better hyper-parameter the MD loss function outperforms the current state of the art (CCS), along with all other loss functions that we have tested.
\end{itemize}

\section{Background}

Contrast-Consistent Search (CCS), developed by \citet{ogpaper}, is an unsupervised method for extracting knowledge from the hidden states of large language models. Given only unlabelled model activations, CCS is able to accurately classify statements according to their truth value. It does this by utilising the negation consistency property of truth: a statement and its negation must have opposite truth values. Framed in terms of probabilities, given the probability $p$ that a proposition is true, the probability its negation is true is $1-p$. CCS works by finding a direction in activation space that satisfies this consistency constraint. 

\textbf{Dataset.} We take a dataset of \emph{contrast pairs} $\{(x^+_i,x^-_i)\}_{i=1}^n$ consisting of natural language statements $x^+_i$ and their logical opposites $x^-_i$. The pairs are formed by taking a question $q_i$ and appending with one of two mutually exclusive answers. Contrast pairs are fed to a pre-trained language model to obtain a set of representations $\{(\phi^+_i,\phi^-_i)\}_{i=1}^n$, where $\phi^{\pm}_i := \phi(x^\pm_i)\in \mathbb{R}^d$ is the activation vector of a particular layer for the input $x^\pm_i$. These representations are then used to train a linear classifier according to an objective designed to enforce negation consistency on contrast pairs.

Below is an example of a contrast pair:
  \begin{itemize}
    \item[] $q_i$\textit{: ``Are cats mammals?''}
    \item[] $x_i^-$\textit{: ``Are cats mammals? No.''}
    \item[] $x_i^+$\textit{: ``Are cats mammals? Yes.''}
\end{itemize}

In order to avoid training the classifier to simply detect the presence of the mutually exclusive answers, the sets $\{ \phi^+_i\}$ and $\{ \phi^-_i\}$ should be independently normalized. The normalized representations are given by
\begin{equation}
    \tilde{\phi}^\pm_i := \frac{\phi^\pm_i - \mu^{\pm}}{\sigma^{\pm}},
\end{equation}
where $(\mu^{\pm}, \sigma^{\pm})$ are the means and standard deviations of the respective sets. For convenience, we will omit the tilde and simply use $\phi_i^{\pm}$ to represent the normalized representations.

\textbf{Loss Function.} A linear classifier $p_{\theta, b}:\phi \to \sigma ( \theta^{T} \phi + b)$ is trained on the normalized activations, where $\sigma$ is the sigmoid function, $\theta$ is a vector of weights and $b$ is a bias. The loss function is given by
\begin{equation}
    L_{\text{CCS}}(\theta,b) := \frac 1 n \sum_{i=1} ^n ~\left[1-p_{\theta, b}(\phi^+_i) - p_{\theta, b}(\phi^-_i)\right]^2 + \min \left\{ p_{\theta, b}(\phi^+_i), p_{\theta, b}(\phi^-_i)\right\}^2.
\end{equation}
The first term encourages the classifier to find features that are negation-consistent and the second term is included to disincentivise the degenerate assignment $p_{\theta, b}(\phi^+_i) = p_{\theta, b}(\phi^-_i) = 0.5$.

\textbf{Metrics.} The accuracy of CCS is calculated using the ground-truth dataset labels. That is, CCS probers are trained in an unsupervised way, but evaluated using supervised labels.

\textbf{Inference.} To make a prediction on an example $q_i$, after training, the average 
\begin{equation} \label{eq prob}
    \bar{p}_i(q_i):=\frac{1}{2}(p_{\theta, b}(\phi^+_i) +(1 - p_{\theta, b}(\phi^-_i)))
\end{equation}
is compared to $0.5$ with $\bar{p_i}>0.5$ corresponding to either the answer "yes" or "no" based on whichever gives the maximum accuracy on a given test set.\footnote{Note that this supervised method is used for simplicity of evaluation; \citet{ogpaper} describe a completely unsupervised method based on conjunctions for determining this assignment.}

\citet{ogpaper} show that a CCS prober trained with the above loss function outperforms zero-shot outputs of the model with a mean accuracy of 71.2\%, against 67.2\% for zero-shot.

\section{Methods: Introducing New Loss Functions}

In this section we introduce an alternative loss function to CCS. First, we provide some conceptual clarification and address a possible misconception about how CCS works. The code used to run our experiments can be found at \url{https://github.com/ash-ai-safety-hub/g3-nandi}.

\subsection{Clarifying CCS}\label{sec:clarifying-CCS}
A natural guess for how CCS is able to accurately classify truth is that the normalized model representations are approximately clustered according to truth and that CCS is able to find a hyperplane that separates these two clusters. However, this explanation turns out to be incorrect. 
In Appendix \ref{app:no-hyperplane} we provide a specific example showing that model activations do not in fact cluster in this way.
Recall that after training, an example $q_i$ is classified according to $\bar{p}(q_i)>0.5$. Using equation \ref{eq prob}, this condition reduces to
\begin{align*}
    \sigma ( \theta^{T} \phi^+_i + b)&>\sigma ( \theta^{T} \phi^-_i + b) \\
    \Rightarrow \theta^T(\phi^+_i-\phi^-_i)&>0.
\end{align*}
We see that CCS is classifying only according to the displacement vectors $(\phi^+_i-\phi^-_i)$. Since these are translation invariant, CCS does not require the contrast pairs to be separable by a hyperplane. 

Additionally, the original paper presents CCS as learning probabilities for the truth values of contrast pairs. We suggest abandoning this probabilities framing. We show in Appendix \ref{app:histogram} that CCS can still perform well even when the probabilities are strongly clustered around 0.5. This paints a different picture to that presented in \citep{ogpaper}.

\subsection{Midpoint-Displacement (MD) Loss Function}\label{sec:MD}
In this section we present a heuristic explanation of CCS's optimization target and use this explanation to introduce a new loss function. Note first that the CCS loss function incentivises increasing the separation of the prober outputs of contrast pairs $|p( \phi_{i}^{+})-p(\phi_{i}^{-})|$. 
Consider a CCS prober $p ( \phi ) = \sigma ( \theta^{T} \phi + b)$ in which $\theta$ is constrained to a fixed norm\footnote{We use $|...|$ to denote the Euclidean norm throughout this paper.} $|\theta| = c$.

Using the normalized weight vector $\hat{\theta}$, we define the following quantities:
\begin{equation} \label{eq}
\begin{split}
    u_{i} &:= \phi_{i}^{+} - \phi_{i}^{-}, \text{ and}\\
    \sigma_d^{2} &:= \frac{1}{n} \sum_{i} (\hat{\theta}^{T} u_{i})^{2}.
\end{split}
\end{equation}
Here $u_i$ is the displacement between the activations of a contrast pair and 
$\sigma_d^{2}$ is the mean square separation of the activations of the contrast pairs along the direction $\theta$.

Furthermore, we analogously define
\begin{equation} \label{eq}
\begin{split}
    v_{i} &:= \phi_{i}^{+}+ \phi_{i}^{-}, \text{ and} \\
    \sigma_m^{2} &:= \frac{1}{n} \sum_{i} (\hat{\theta}^{T} v_{i})^{2}.
\end{split}
\end{equation}
Here $\dfrac{v_{i}}{2}$ is the midpoint of the activations of a contrast pair and $\dfrac{\sigma_m^{2}}{4}$ is the mean square value of the midpoint of the activations of the contrast pairs along the direction \( \theta \).

In order for CCS to increase the difference between prober outputs, one might expect that CCS finds a direction that increases the difference of the prober inputs (since sigmoid is a monotonically increasing function). That is to say, one might expect CCS will find a direction that maximises $\sigma_d^{2}$.

However, if $\sigma_{m}^{2}$ is much larger than $\sigma_d^{2}$, then the input to the sigmoid for each contrast pair (i.e. $\theta^{T} \phi_{i}^{+} + b$ and $\theta^{T} \phi_{i}^{-} + b$) will be pushed into the same saturation regime of the sigmoid. This results in a lower difference in prober outputs of contrast pairs, which in turn results in a trade off between maximising $\sigma_d^{2}$ while minimising $\sigma_{m}^{2}$.

It should be stressed that this trade-off between $\sigma_{d}^{2}$ and $\sigma_{m}^{2}$ is purely an artifact of the double saturation of sigmoid used in the CCS prober.
Since this trade off should occur no matter what $|\theta|=c$ is constrained to, we propose that the unconstrained CCS prober is in general optimising for some balance between $\sigma_{d}^{2}$ and $\sigma_{m}^{2}$.

To test this hypothesis, we propose a new loss function and demonstrate that this new loss function is a good proxy optimisation target for CCS. The new loss function is given by
\begin{equation}
    L_{\text{MD}} = (\lambda - 1) \sigma^2_d + \lambda \cdot \sigma^2_m, 
\end{equation}
where $\lambda \in [0,1]$ is a hyper-parameter controlling the relative trade off between $\sigma_{d}^{2}$ and $\sigma_{m}^{2}$, and the weight vector $\theta$ is constrained to satisfy $|\theta|=1$.

In Appendix \ref{app:mean-loss} we introduce two new loss functions: the Mean Absolute (MA) loss function and the Square Mean Root (SMR) loss function. In the following experiments we compare the MD loss function with a number of other loss functions, including the CCS, MA and SMR loss functions along with Principal Component Analysis (PCA).

\section{Results: Comparison of MD with CCS}\label{sec:comparison-MD-CCS}

In this section we investigate the empirical similarity between the MD and CCS loss functions.
The MD-CCS and MD-Accuracy (MD-Acc) loss functions are both trained using the Midpoint-Displacement loss function but with different hyper-parameter searches.
The implementation details of our experiments can be found in Appendix \ref{app:implementation}.

For comparison, we include probers trained using a variety of loss functions.
The MA and SMR loss functions are both based on taking the mean displacements (further details can be found in Appendix \ref{app:mean-loss}).
The PCA loss function identifies the first principal component of the displacements. The random probers (Rand.) are found by randomly initializing 10 weight vectors and taking the average resulting accuracy. Lastly, the supervised probers (Superv.) are probers trained on labelled data with the same structure as the CCS probers, $p(\phi) = \sigma (\theta^{T} \phi +b)$.

In Table \ref{tab:acc-main} we show the test accuracies of probers trained using the MD loss function on various datasets and models. We tentatively find that the accuracies of the new probers are similar to those achieved by CCS, and often out-perform CCS.
Note that MD-Acc probers get a higher test accuracy than both the MD-CCS and CCS probers for three out of four models, for an average difference of around 4\%.

% discuss why supervised has high cosine similarity.

% accuracies
\begin{table*}[h!]
\begin{center}
\begin{tabular}{ || c || c | c | c | c | c | c | c | c ||} 
\hline
 \multirow{2}{*}{Model} & \multicolumn{8}{c||}{Loss Function} \\
 \cline{2-9}
  & CCS & MD-CCS & MD-Acc & MA & SMR & PCA & Rand. & Superv. \\
 \hline
 \hline
 UQA (E) & 0.6863 & 0.6902 & 0.7414 & 0.7399 & \textbf{0.7419} & 0.7383 & 0.6363 & 0.8839 \\
 \hline
 UQA (D) & \textbf{0.8305} & 0.8200 & 0.8180 & 0.7550 & 0.7460 & 0.7525 & 0.6286 & 0.9140 \\
 \hline
 DeBERTa & 0.7740 & 0.7855 & \textbf{0.8735} & 0.8650 & 0.8585 & 0.8605 & 0.7288 & 0.9135 \\
 \hline
 GPT-Neo & 0.5510 & 0.5755 & \textbf{0.5898} & 0.5820 & 0.5555 & 0.5737 & 0.5603 & 0.7580 \\
 \hline
 \hline
 Average & 0.7105 & 0.7178 & \textbf{0.7557} & 0.7355 & 0.7255 & 0.7313 & 0.6385 & 0.8674 \\
 \hline
\end{tabular}
%\vspace{0.2cm}
\caption{We compare test accuracies of different loss functions averaged over five datasets, using the activations of a number of models. For each row we have emboldened the loss function that obtained the highest average test accuracy, not including the supervised loss. The (E) and (D) labels refer to the encoder and decoder layers of the UQA model.}
\label{tab:acc-main}
\end{center}
\end{table*}

% cosine similarity
\begin{table*}[h!]
\begin{center}
\begin{tabular}{ || c || c | c | c | c | c | c | c | c || } 
\hline
 \multirow{2}{*}{Model} & \multicolumn{8}{c||}{Loss Function} \\
 \cline{2-9}
  & CCS & MD-CCS & MD-Acc & MA & SMR & PCA & Rand. & Superv. \\
 \hline
 \hline
 UQA (E) & 0.8359 & \textbf{0.7034} & 0.2995 & 0.1448 & 0.1991 & 0.2406 & 0.0222 & 0.2583 \\
 \hline
 UQA (D) & 0.8787 & \textbf{0.7269} & 0.5303 & 0.1687 & 0.2432 & 0.1792 & 0.0228 & 0.6014 \\
 \hline
 DeBERTa & 0.8643 & \textbf{0.6209} & 0.2786 & 0.2309 & 0.2024 & 0.0741 & 0.0202 & 0.4617 \\
 \hline
 GPT-Neo & 0.5277 & \textbf{0.4830} & 0.4164 & 0.0226 & 0.0485 & 0.1901 & 0.0245 & 0.1347 \\
 \hline
 \hline
 Average & 0.7767 & \textbf{0.6336} & 0.3812 & 0.1418 & 0.1733 & 0.1710 & 0.0224 & 0.3640 \\
 \hline
\end{tabular}
%\vspace{0.2cm}
\caption{We compute the average cosine similarities of the directions found using different loss functions to the directions of 20 CCS probers. We average over five datasets using the activations of four different models. For each row we have emboldened the loss function that obtained the highest average cosine similarity with CCS, not including the CCS loss function. Note that the (E) and (D) refer to the encoder and decoder layers of the UQA model.}
\label{tab:cosine-main}
\end{center}
\end{table*}

In Table \ref{tab:cosine-main} we find that the average cosine similarity between the weight vector of the CCS prober and the weight vector of the prober trained using our new MD method is about 0.63. For reference, the probability of two uniformly sampled 1024-dimensional unit vectors having a cosine similarity of 0.63 or higher is approximately $10^{-237}$. Note that the CCS probers had an average cosine similarity with themselves of only 0.78. This suggests that the MD-CCS loss function is a good proxy optimization target for CCS.

The only difference between the MD-CCS and MD-Acc loss functions is the value of their hyper-parameter $\lambda$ that controls the relative trade off between $\sigma_{d}^{2}$ and $\sigma_{m}^{2}$. 
Since they give very different cosine similarities, we find the similarity of MD to CCS is dependent on this trade off.

In Appendix \ref{app:granular}, we include a breakdown of the experimental results for each dataset and model along with the hyper-parameters that are used for each loss function. It should be noted that the hyper-parameter used for the MD-CCS loss function does not change substantially between datasets or models. This suggests that the form of the proxy optimization target we have identified for CCS is robust to changes in the dataset and model.

\section{Discussion}

Our Midpoint-Displacement loss function sheds light on which aspects of the data CCS is picking up on.
We find that the Midpoint-Displacement loss function produces probers that behave very similarly to those produced by CCS, as measured by cosine similarity.
We verified that other reasonable loss functions, with comparable accuracy to CCS, do not get high cosine similarity to CCS.

Our findings suggest that the specific loss function formulation is not essential for performance, in that there are multiple loss functions that achieve high accuracy. 
Instead, the unique training data that CCS uses, which allows the identification of displacement between $\phi^+$ and $\phi^-$, is what drives the success of CCS.

\subsection{Limitations}

While the new loss functions we have proposed are unsupervised, they contain a hyper-parameter. The hyper-parameter is chosen using a supervised grid search and therefore supervised labels are currently required to use our loss functions. Additionally, the results obtained in this paper could be strengthened by testing across more datasets and models.

\subsection{Future Work}

Future work could also consider developing an unsupervised method of determining the hyper-parameter based on data statistics. This would eliminate the need for any supervised labels. Alternatively, we could establish a hyper-parameter with high transferability between datasets. 
Moreover, we propose that a small number of supervised examples may be sufficient to find a dataset-specific hyper-parameter.
% We could also investigate whether a small number of supervised examples is sufficient to find a dataset-specific hyper-parameter.

In this paper we have evaluated similarity between probers through the metric of cosine similarity. Future work could additionally consider comparing the behavioral similarity between probers, by measuring their empirical pairwise agreement for contrast pairs in the test dataset. 

% \newpage

% \section*{Social Impact Statement}

\subsection{Social Impact Statement}

% Eliciting Latent Knowledge (ELK) is the field of study that aims to extract the internal knowledge of language models. It is hoped that work within ELK will help detect lying, deception and hallucination. Detecting lying and deception are thought to be important steps towards aligning advanced AI systems, and therefore mitigating the risks of an AI catastrophe. Detecting hallucination is important to prevent the spread of misinformation. CCS is a nascent method of extracting truthfulness from language models. 
% This paper provides clarifications on how CCS is able to extract information from the activations of a language model. We hope our results are able to inform work on CCS and future evaluation techniques that aim to extract latent knowledge. We do not foresee harmful applications of our work.

% To detect deception in LLMs, we may compare model outputs to the model’s latent knowledge. Uncovering this latent knowledge involves extracting knowledge from unlabeled hidden representations. CCS is a nascent method for extracting truthfulness from language models.
% This paper aims to clarify CCS and we hope our results are able to improve CCS and inform future evaluation techniques that aim to extract latent knowledge. 
% We do not foresee harmful applications of our work.

Our work is motivated by avoiding harmful behaviors of language models, such as deception.
To detect deception in LLMs, we may compare model outputs to the model’s latent knowledge.  
CCS is a nascent method for extracting truthfulness from latent representations.
We aim to clarify CCS and we hope our results will inform future evaluation techniques that extract latent knowledge. 
We do not foresee harmful applications of our work.

\subsection*{Acknowledgments}

This research was supported by the AI Safety Hub Labs programme.

%% file: write-up-sections/appendix-opt.tex
\appendix

\section{Clarifying CCS}

\subsection{CCS Does Not Determine Truth Using a Hyperplane}\label{app:no-hyperplane}

Let $\theta$ be the weight vector of the linear prober found by CCS and denote the unit-normalised weight vector by $\hat{\theta}$.
We call this normalised weight vector `the direction found by CCS'.

\begin{figure}[h]
    \centering
    \includegraphics[width=0.95\linewidth]{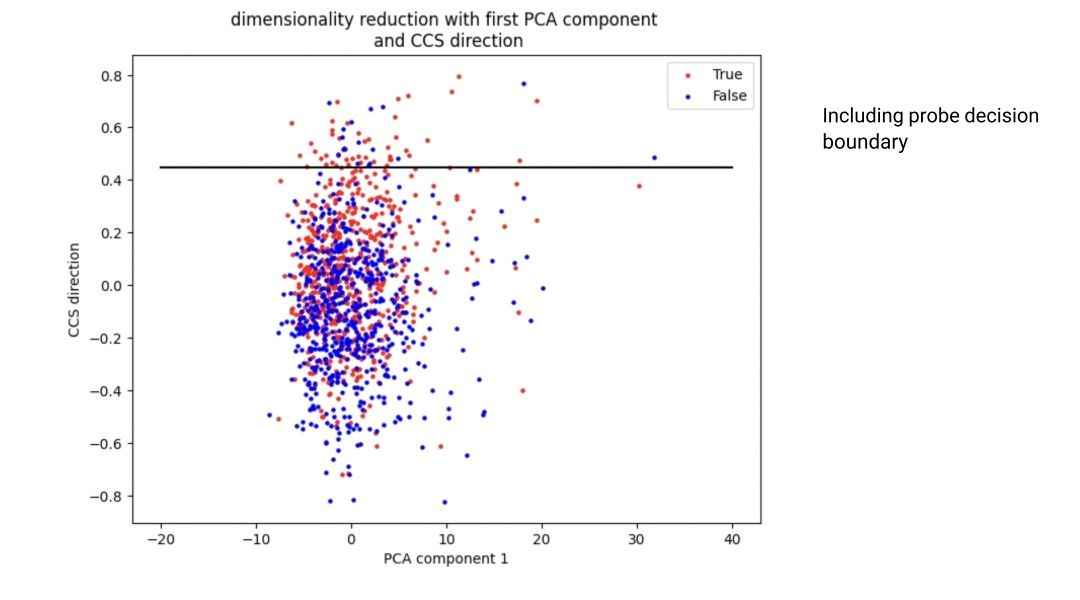}
    \caption{
    We consider activations of the T5-base model on the BoolQ train dataset.
    On the $x$-axis we plot the projections of the activations $\phi_i$ onto the first principle component and on the $y$-axis we plot the datapoints $\phi_i$ projected onto $\hat{\theta}$.
    We colour the datapoints $\phi_i$ by the (ground-truth) truth-labels `true' (red) and `false' (blue) of the datapoints $x_i$.
    The horizontal line indicates the inputs for which CCS outputs 0.5.}
    \label{fig:no-hyperplane}
\end{figure}

In Figure \ref{fig:no-hyperplane} we project the activations vectors $\phi_i$ corresponding to datapoints $x_i$ onto the first principal component and onto the direction found by CCS (when trained on the decoder). 
We consider the activations for which CCS outputs 0.5 and show that this `decision boundary' does not cleanly separate the true and false datapoints.

\subsection{CCS is Miscalibrated}\label{app:histogram}
\begin{figure}[h]
\centering
\begin{subfigure}{.5\textwidth}
    \centering
    \includegraphics[width=0.95\linewidth]{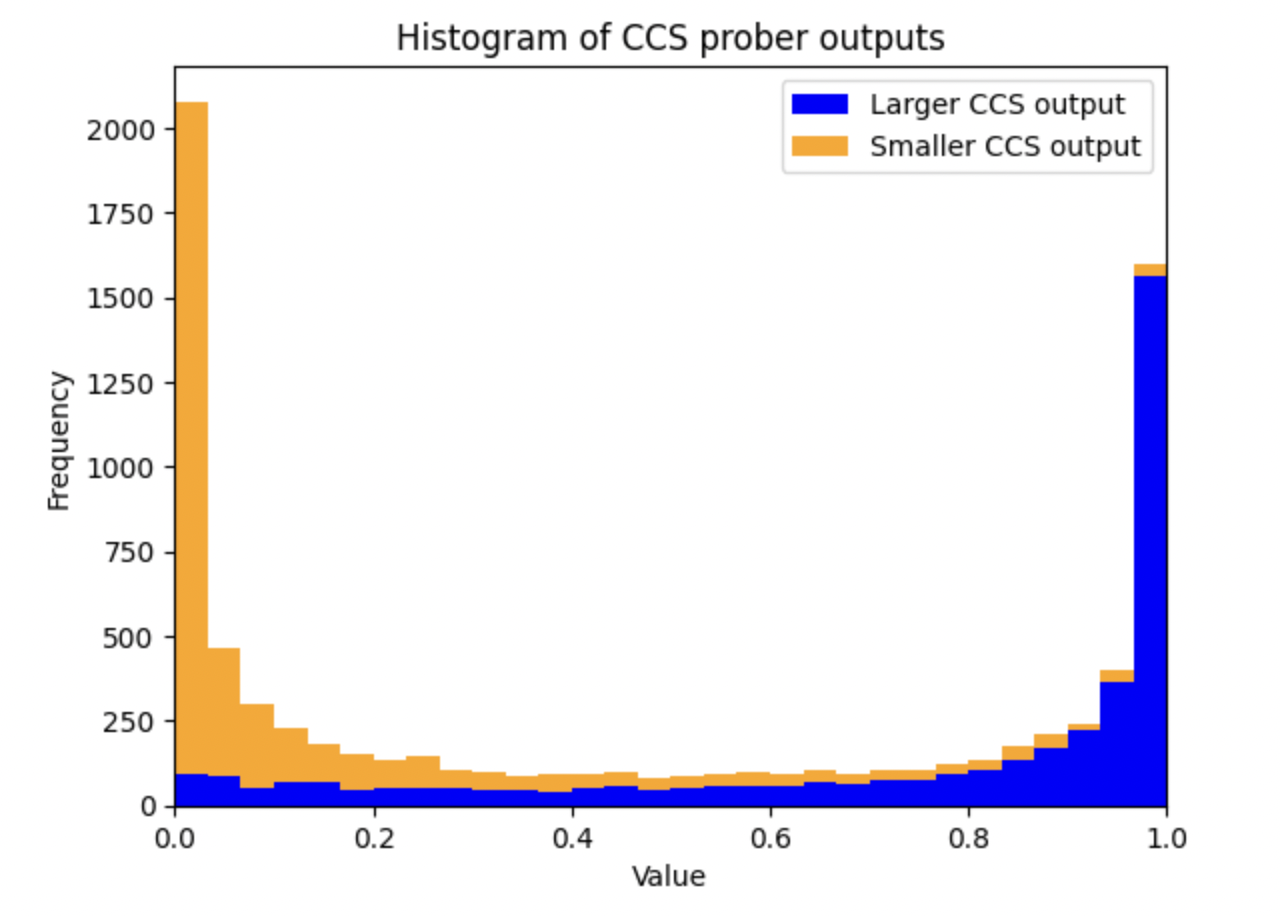}
    \caption{}
    \label{fig:hist-enc}
\end{subfigure}%
\begin{subfigure}{.5\textwidth}
    \centering
    \includegraphics[width=0.95\linewidth]{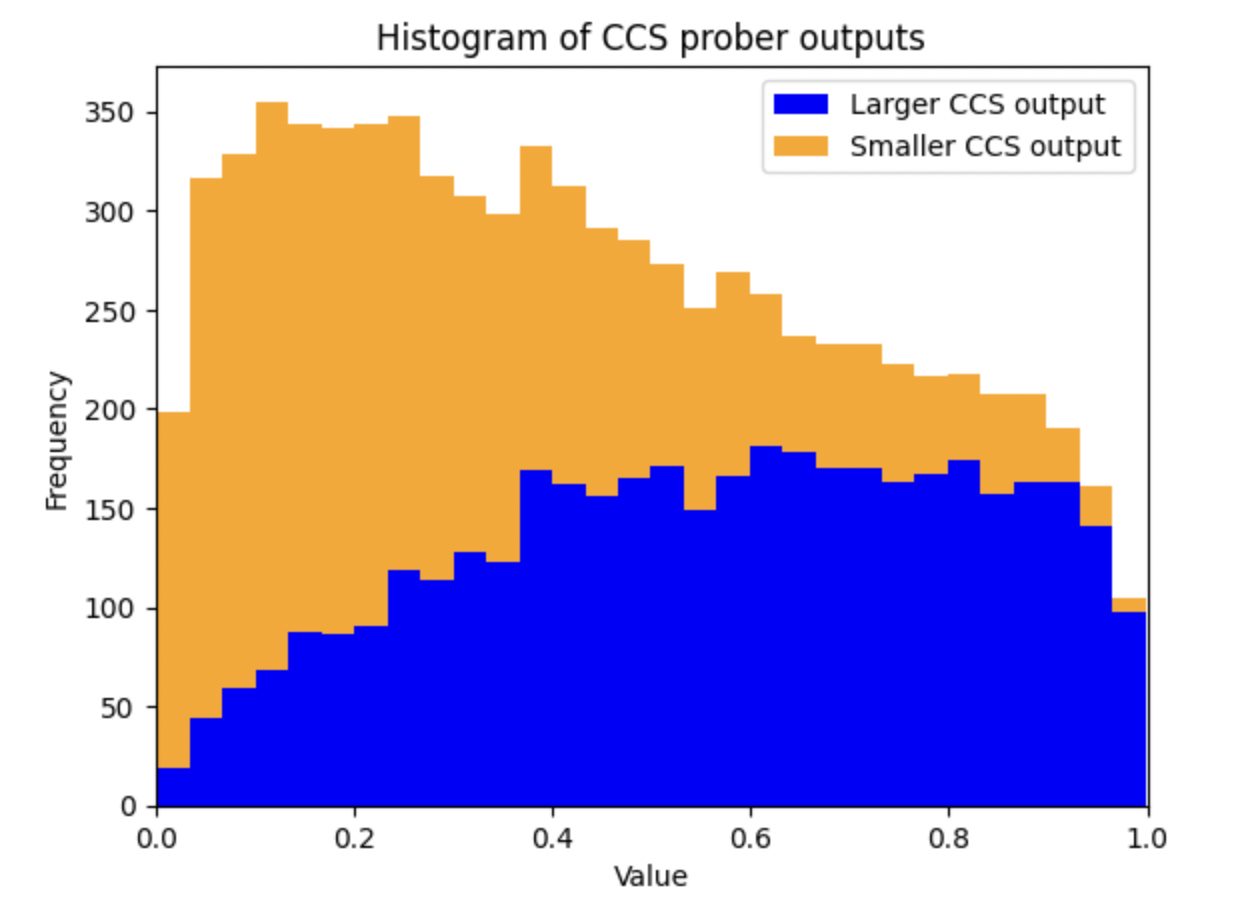}
    \caption{}
    \label{fig:hist-dec}
\end{subfigure}
\caption{(a) Histogram displaying the CCS prober outputs evaluated on the last hidden state of the encoder of UnifiedQA T5-Large for the BoolQ dataset. (b) Histogram displaying the CCS prober outputs evaluated on the last hidden state of the decoder of UnifiedQA T5-Large for the BoolQ dtatset. Despite the encoder having a higher confidence in the prober outputs, the encoder has a lower test accuracy (0.523) than the decoder (0.978).}
\label{fig:test}
\end{figure}

In Figure \ref{fig:test}, we show histograms of CCS prober outputs evaluated on the last hidden state of both the encoder and decoder of UQA for the BoolQ dataset (compare to Figure 1 of \citet{ogpaper}). Interpreting the output of CCS as a probability would suggest much higher confidence for the encoder than the decoder. However, from Appendix \ref{app:granular}, the test accuracy on the encoder (0.523) is significantly worse than on the decoder (0.978). This suggests that the ouput of CCS should not be interpreted as probabilities.  

% \citep{ogpaper} visualize the effect of their loss function on the confidence of a trained prober in the histogram Figure 1. of their paper. 
% This histogram looks very similar to the histogram in Figure \ref{fig:hist-enc}. 

% Note that in Table \ref{tab:acc-main} we see that UnifiedQA accuracy for the decoder is much higher than for the encoder. 
% However, it is also the decoder in Figure \ref{fig:hist-dec} which looks very different from a calibrated probability distribution.

\newpage
\section{Mean-Based Loss Functions}\label{app:mean-loss}
%Maybe move this section to the appendix?

% Discuss connection to MD loss.

We define the displacement vector between the activations of a contrast pair as follows.
\begin{align*}
     u_{i} = \phi^{+}_{i} - \phi^{-}_{i}.
\end{align*}
We define the mean and standard deviation of the modulus of the displacements along a direction $\theta$ as follows:
\begin{align*}
     \mu &= \frac1n \sum_{i} |\theta^{T} u_i | ,\\
     \sigma &= \sqrt{ \frac1n \sum_{i} (| \theta^{T} u_i | - \mu)^2 }.
\end{align*}
We then define the Mean Absolute (MA) loss function as 
\begin{align*}
     L_{\text{MA}} = (1-\lambda) \mu + \lambda \cdot \sigma.
\end{align*}
The Square Mean Root (SMR) is given by
\begin{align*}
     \mu_{\text{SMR}} &= \sqrt{\frac1n \sum_{i} ( \theta^{T} u_i )^2} .
\end{align*}
Analogously, we define the SMR loss function to be
\begin{align*}
     L_{\text{SMR}} = (1 - \lambda) \mu_{\text{SMR}} + \lambda \cdot \sigma.
\end{align*}
In both loss functions, we constrain $|\theta|=1$.

\section{Implementation Details}\label{app:implementation}

For the experiments presented in this paper, all probers were trained for 1000 epochs with a learning rate of 0.01. All of the experiments were run for the following five datasets:

\begin{itemize}
    \item IMDB~\citep{maas-etal-2011-learning}
    \item Amazon~\citep{amazon}
    \item BoolQ~\citep{boolq}
    \item RTE~\citep{superglue}
    \item AG News~\citep{agnews}
\end{itemize}

and the following three models:
\begin{itemize}
    \item UnifiedQA T5-Large (encoder-decoder architecture) ~\citep{khashabi2020unifiedqa}
    \item GPT-Neo (decoder-only architecture) ~\citep{gpt-neo}
    \item DeBERTa (encoder-only architecture) ~\citep{he2021deBERTa}
\end{itemize}
We use the hidden states of the encoder and decoder where available. In our experiments, we used 1000 examples from each dataset consisting of a 600-400 train-test split. For each loss function on each dataset and model, we use a grid search to determine the hyper-parameter that maximised train accuracy as follows:

\begin{enumerate}
    \item Create eleven hyper-parameter values by splitting the interval $[0,0.99]$ into eleven evenly spaced points (separated by 0.099).
    \item For each of these values, we train three probers with a random seed. We evaluate the average train accuracy over the three probers.
    \item We identify the hyper-parameter with the highest average train accuracy. Let this hyper-parameter be $\lambda^{\star}$.
    \item Consider the new interval given by $[\lambda^{\star}-0.099,\lambda^{\star}+0.099] \cap [0,0.99]$.
    \item Repeat the process again on the new interval by splitting it evenly into eleven points.
    \item Identify the hyper-parameter with the highest average train accuracy for the points on this new interval. This should calculate the hyper-parameter with a precision of roughly $\pm 0.02$.
\end{enumerate}

An analogous grid search is performed to identify the hyper-parameter used for the MD-CCS loss function, by finding the hyper-parameter that maximises average cosine similarity with 20 CCS probers. In the case of the MD-CCS hyper-parameter, the initial interval we consider is $[0.9, 0.999]$ as opposed to $[0,0.99]$ (initial experiments always converged to a hyper-parameter in this interval - so we used this initial interval to reduce compute time).

Having found the optimal hyper-parameter $\lambda^{\star}$, we then train ten probers with $\lambda = \lambda^{\star}$ and random seeds. From these ten trained probers, we pick the prober that minimises the loss function (this is in line with the method used in \cite{ogpaper}). This is the prober that is used to compute the values found in the tables of this paper.

\newpage
\section{Granular Data for Models and Specific Datasets} \label{app:granular}

\subsection{Accuracies}

% accuracies
\begin{table}[h!]
\begin{center}
\begin{tabular}{ || c || c | c | c | c | c | c | c | c ||} 
\hline
 \multirow{2}{*}{Dataset} & \multicolumn{8}{c||}{Loss Function} \\
 \cline{2-9}
  & CCS & MD-CCS & MD-Acc & MA & SMR & PCA & Rand. & Superv. \\
 \hline
 \hline
 IMDB & 0.9200 & 0.9225 & 0.9200 & 0.9225 & 0.9200 & 0.9220 & 0.7908 & 0.9300 \\
 \hline
 Amazon & 0.9438 & 0.9469 & 0.9469 & 0.9469 & 0.9469 & 0.9469 & 0.7344 & 0.9469 \\
 \hline
 BoolQ & 0.5225 & 0.5167 & 0.7825 & 0.7900 & 0.7925 & 0.7800 & 0.6047 & 0.8875 \\
 \hline
 AG News & 0.5275 & 0.5375 & 0.5300 & 0.5275 & 0.5275 & 0.5150 & 0.5152 & 0.9500 \\
 \hline
 RTE & 0.5175 & 0.5275 & 0.5275 & 0.5125 & 0.5225 & 0.5275 & 0.5365 & 0.7050 \\
 \hline
 \hline
 Average & 0.6863 & 0.6902 & 0.7414 & 0.7399 & 0.7419 & 0.7383 & 0.6363 & 0.8839 \\
 \hline
\end{tabular}
\vspace{0.1cm}
\caption{We compare test accuracies of different loss functions using the last hidden state of the encoder of UnifiedQA T5-Large for a number of datasets.}
\label{tab:acc}
\end{center}
\end{table}

\vspace{-0.3cm}

% accuracies
\begin{table}[h!]
\begin{center}
\begin{tabular}{ || c || c | c | c | c | c | c | c | c ||} 
\hline
 \multirow{2}{*}{Dataset} & \multicolumn{8}{c||}{Loss Function} \\
 \cline{2-9}
  & CCS & MD-CCS & MD-Acc & MA & SMR & PCA & Rand. & Superv. \\
 \hline
 \hline
 IMDB & 0.9275 & 0.9250 & 0.9200 & 0.9325 & 0.9325 & 0.9150 & 0.7190 & 0.9200 \\
 \hline
 Amazon & 0.9400 & 0.9425 & 0.9450 & 0.6050 & 0.5900 & 0.7675 & 0.6355 & 0.9425 \\
 \hline
 BoolQ & 0.9775 & 0.9625 & 0.9625 & 0.9800 & 0.9875 & 0.8725 & 0.6303 & 0.9850 \\
 \hline
 AG News & 0.5924 & 0.5300 & 0.5300 & 0.5675 & 0.5325 & 0.5175 & 0.5750 & 0.9550 \\
 \hline
 RTE & 0.7150 & 0.7400 & 0.7325 & 0.6900 & 0.6875 & 0.6900 & 0.5833 & 0.7675 \\
 \hline
 \hline
 Average & 0.8305 & 0.8200 & 0.8180 & 0.7550 & 0.7460 & 0.7525 & 0.6286 & 0.9140 \\
 \hline
\end{tabular}
\vspace{0.1cm}
\caption{We compare test accuracies of different loss functions using the last hidden state of the decoder of UnifiedQA T5-Large for a number of datasets.}
\label{tab:acc}
\end{center}
\end{table}

\vspace{-0.3cm}

% accuracies
\begin{table}[h!]
\begin{center}
\begin{tabular}{ || c || c | c | c | c | c | c | c | c ||} 
\hline
 \multirow{2}{*}{Dataset} & \multicolumn{8}{c||}{Loss Function} \\
 \cline{2-9}
  & CCS & MD-CCS & MD-Acc & MA & SMR & PCA & Rand. & Superv. \\
 \hline
 \hline
 IMDB & 0.6225 & 0.7050 & 0.7650 & 0.7400 & 0.5225 & 0.7435 & 0.5525 & 0.8225 \\
 \hline
 Amazon & 0.5950 & 0.6200 & 0.6200 & 0.6175 & 0.7100 & 0.5725 & 0.6422 & 0.8800 \\
 \hline
 BoolQ & 0.5075 & 0.5350 & 0.53375 & 0.5075 & 0.5075 & 0.5225 & 0.5360 & 0.5800 \\
 \hline
 AG News & 0.5250 & 0.5075 & 0.5150 & 0.5375 & 0.5325 & 0.5175 & 0.5478 & 0.9500 \\
 \hline
 RTE & 0.5050 & 0.5100 & 0.5150 & 0.5075 & 0.5050 & 0.5125 & 0.5232 & 0.5575 \\
 \hline
 \hline
 Average & 0.5510 & 0.5755 & 0.5898 & 0.5820 & 0.5555 & 0.5737 & 0.5603 & 0.7580 \\
 \hline
\end{tabular}
\vspace{0.1cm}
\caption{We compare test accuracies of different loss functions using the last hidden state of the decoder of GTP-Neo for a number of datasets.}
\label{tab:acc}
\end{center}
\end{table}

\vspace{-0.3cm}

% accuracies
\begin{table}[h!]
\begin{center}
\begin{tabular}{ || c || c | c | c | c | c | c | c | c ||} 
\hline
 \multirow{2}{*}{Dataset} & \multicolumn{8}{c||}{Loss Function} \\
 \cline{2-9}
  & CCS & MD-CCS & MD-Acc & MA & SMR & PCA & Rand. & Superv. \\
 \hline
 \hline
 IMDB & 0.9550 & 0.9500 & 0.9475 & 0.9550 & 0.9475 & 0.9500 & 0.8067 & 0.9550 \\
 \hline
 Amazon & 0.9475 & 0.9425 & 0.9475 & 0.9400 & 0.9500 & 0.9425 & 0.8145 & 0.9500 \\
 \hline
 BoolQ & 0.6600 & 0.7475 & 0.8100 & 0.8125 & 0.8125 & 0.8100 & 0.7100 & 0.8275 \\
 \hline
 AG News & 0.5000 & 0.5000 & 0.8250 & 0.7700 & 0.7300 & 0.8250 & 0.6545 & 0.9425 \\
 \hline
 RTE & 0.8075 & 0.7875 & 0.8375 & 0.8475 & 0.8525 & 0.775 & 0.6585 & 0.8925 \\
 \hline
 \hline
 Average & 0.7740 & 0.7855 & 0.8735 & 0.8650 & 0.8585 & 0.8605 & 0.7288 & 0.9135 \\
 \hline
\end{tabular}
\vspace{0.1cm}
\caption{We compare test accuracies of different loss functions using the last hidden state of the encoder of DeBERTa for a number of datasets.}
\label{tab:acc}
\end{center}
\end{table}

\newpage
\subsection{Cosine Similarities}

% cosine similarity
\begin{table}[h!]
\begin{center}
\begin{tabular}{ || c || c | c | c | c | c | c | c | c || } 
\hline
 \multirow{2}{*}{Dataset} & \multicolumn{8}{c||}{Loss Function} \\
 \cline{2-9}
  & CCS & MD-CCS & MD-Acc & MA & SMR & PCA & Rand. & Superv. \\
 \hline
 \hline
 IMDB & 0.9305 & 0.7065 & 0.2929 & 0.2064 & 0.3167 & 0.2930 & 0.0214 & 0.3862 \\
 \hline
 Amazon & 0.9635 & 0.8177 & 0.3149 & 0.2777 & 0.3867 & 0.3150 & 0.0231 & 0.7639 \\
 \hline
 BoolQ & 0.8082 & 0.8000 & 0.0187 & 0.0332 & 0.0442 & 0.0194 & 0.0181 & 0.0860 \\
 \hline
 AG News & 0.5081 & 0.4817 & 0.4758 & 0.1997 & 0.2009 & 0.1835 & 0.0267 & 0.0328 \\
 \hline
 RTE & 0.9691 & 0.7111 & 0.3954 & 0.0069 & 0.0468 & 0.3919 & 0.0215 & 0.0224 \\
 \hline
 \hline
 Average & 0.8359 & 0.7034 & 0.2995 & 0.1448 & 0.1991 & 0.2406 & 0.0222 & 0.2583 \\
 \hline
\end{tabular}
\vspace{0.1cm}
\caption{We compute the average cosine similarities of the directions found using different loss functions to the directions of 20 CCS probers. We use the last hidden state of the encoder of UnifiedQA T5-Large for a number of datasets.}
\label{tab:cosine}
\end{center}
\end{table}

\vspace{-0.8cm}

\begin{table}[h!]
\begin{center}
\begin{tabular}{ || c || c | c | c | c | c | c | c | c || } 
\hline
 \multirow{2}{*}{Dataset} & \multicolumn{8}{c||}{Loss Function} \\
 \cline{2-9}
  & CCS & MD-CCS & MD-Acc & MA & SMR & PCA & Rand. & Superv. \\
 \hline
 \hline
 IMDB & 0.9905 & 0.7111 & 0.4048 & 0.3867 & 0.3815 & 0.2378 & 0.0212 & 0.6790 \\
 \hline
 Amazon & 0.8357 & 0.6021 & 0.3969 & 0.0387 & 0.0076 & 0.0760 & 0.0165 & 0.7499 \\
 \hline
 BoolQ & 0.9850 & 0.8694 & 0.5393 & 0.2768 & 0.2808 & 0.1158 & 0.0194 & 0.8640 \\
 \hline
 AG News & 0.5924 & 0.5348 & 0.4547 & 0.0109 & 0.4274 & 0.3548 & 0.0254 & 0.0219 \\
 \hline
 RTE & 0.9899 & 0.9169 & 0.8558 & 0.1303 & 0.1187 & 0.1115 & 0.0314 & 0.6924 \\
 \hline
 \hline
 Average & 0.8787 & 0.7269 & 0.5303 & 0.1687 & 0.2432 & 0.1792 & 0.0228 & 0.6014 \\
 \hline
\end{tabular}
\vspace{0.1cm}
\caption{We compute the average cosine similarities of the directions found using different loss functions to the directions of 20 CCS probers. We use the last hidden state of the decoder of UnifiedQA T5-Large for a number of datasets.}
\label{tab:cosine}
\end{center}
\end{table}

\vspace{-0.8cm}

\begin{table}[h!]
\begin{center}
\begin{tabular}{ || c || c | c | c | c | c | c | c | c || } 
\hline
 \multirow{2}{*}{Dataset} & \multicolumn{8}{c||}{Loss Function} \\
 \cline{2-9}
  & CCS & MD-CCS & MD-Acc & MA & SMR & PCA & Rand. & Superv. \\
 \hline
 \hline
 IMDB & 0.6296 & 0.2483 & 0.0785 & 0.0549 & 0.0367 & 0.0624 & 0.0212 & 0.3465 \\
 \hline
 Amazon & 0.3005 & 0.3059 & 0.3066 & 0.0092 & 0.0359 & 0.1351 & 0.0266 & 0.2310 \\
 \hline
 BoolQ & 0.6478 & 0.6683 & 0.6628 & 0.0175 & 0.0175 & 0.3581 & 0.0213 & 0.0136 \\
 \hline
 AG News & 0.3644 & 0.4627 & 0.3123 & 0.0133 & 0.0123 & 0.1781 & 0.0298 & 0.0531 \\
 \hline
 RTE & 0.6964 & 0.7296 & 0.7218 & 0.0183 & 0.1400 & 0.2166 & 0.0237 & 0.0295 \\
 \hline
 \hline
 Average & 0.5277 & 0.4830 & 0.4164 & 0.0226 & 0.0485 & 0.1901 & 0.0245 & 0.1347 \\
 \hline
\end{tabular}
\vspace{0.1cm}
\caption{We compute the average cosine similarities of the directions found using different loss functions to the directions of 20 CCS probers. We use the last hidden state of the decoder of GPT-Neo for a number of datasets.}
\label{tab:cosine}
\end{center}
\end{table}

\begin{table}[h!]
\vspace{-0.8cm}
\begin{center}
\begin{tabular}{ || c || c | c | c | c | c | c | c | c || } 
\hline
 \multirow{2}{*}{Dataset} & \multicolumn{8}{c||}{Loss Function} \\
 \cline{2-9}
  & CCS & MD-CCS & MD-Acc & MA & SMR & PCA & Rand. & Superv. \\
 \hline
 \hline
 IMDB & 0.9376 & 0.5504 & 0.5466 & 0.2599 & 0.1685 & 0.0554 & 0.0143 & 0.6811 \\
 \hline
 Amazon & 0.9538 & 0.2760 & 0.2406 & 0.1657 & 0.2401 & 0.0648 & 0.0178 & 0.8104 \\
 \hline
 BoolQ & 0.7823 & 0.7377 & 0.0357 & 0.0448 & 0.0543 & 0.0358 & 0.0175 & 0.2912 \\
 \hline
 AG News & 0.7003 & 0.6401 & 0.0145 & 0.1750 & 0.0140 & 0.0146 & 0.0222 & 0.0435 \\
 \hline
 RTE & 0.9476 & 0.9002 & 0.5557 & 0.5093 & 0.5349 & 0.1998 & 0.0294 & 0.4823 \\
 \hline
 \hline
 Average & 0.8643 & 0.6209 & 0.2786 & 0.2309 & 0.2024 & 0.0741 & 0.0202 & 0.4617 \\
 \hline
\end{tabular}
\vspace{0.1cm}
\caption{We compute the average cosine similarities of the directions found using different loss functions to the directions of 20 CCS probers. We use the last hidden state of the encoder of DeBERTa for a number of datasets.}
\label{tab:cosine}
\end{center}
\end{table}

\newpage
\subsection{Results from Hyper-Parameter Grid Search}

We find that the MD-CCS hyper-parameter is always close to 1. 
This suggests that the trade-off most similar to CCS is one that prioritises minimising $\sigma_{m}^{2}$ (which has a coefficient of at least 0.9) over maximising $\sigma_{d}^{2}$ (which has a coefficient of at most 0.1).

For the MD-Acc loss function, we find that the hyper-parameter is often either close to 0 or close to 1. 
When the hyper-parameter is close to 1, the MD-CCS and MD-Acc have landed on the same hyper-parameter.
On the other hand, 
when MD-Acc has a $\lambda$ is close to 0,
the loss function is  very similar to PCA.

\begin{table}[h!] 
\begin{center} 
\begin{tabular}{ || c || c | c | c | c ||}  
\hline 
\multirow{2}{*}{Dataset} & \multicolumn{4}{c||}{Loss Function} \\ 
\cline{2-5} 
  & MD-CCS & MD-Acc & MA & SMR \\ 
\hline 
\hline 
IMDB & 0.9891 & 0.0000 & 0.7524 & 0.6930 \\ 
\hline 
Amazon & 0.9812 & 0.0000 & 0.6930 & 0.5940  \\ 
\hline 
BoolQ & 0.9752 & 0.0000 & 0.0000 & 0.0000  \\ 
\hline 
AG News & 0.9871 & 0.9801 & 0.0000 & 0.0000  \\ 
\hline 
RTE & 0.9891 & 0.0990 & 0.9900 & 0.1980  \\ 
\hline 
\end{tabular} 
\vspace{0.1cm}
\caption{Hyper-parameters values  for the encoder of UnifiedQA T5-Large.} 
\label{tab:hyper_parameters_UQA_encoder} 
\end{center} 
\end{table}

\vspace{-0.3cm}

\begin{table}[h!] 
\begin{center} 
\begin{tabular}{ || c || c | c | c | c ||}  
\hline 
\multirow{2}{*}{Dataset} & \multicolumn{4}{c||}{Loss Function} \\ 
\cline{2-5} 
  & MD-CCS & MD-Acc & MA & SMR \\ 
\hline 
\hline 
IMDB & 0.9871 & 0.8910 & 0.6732 & 0.4950 \\ 
\hline 
Amazon & 0.9653 & 0.7128 & 0.6930 & 0.6930  \\ 
\hline 
BoolQ & 0.9535 & 0.7920 & 0.5940 & 0.5544  \\ 
\hline 
AG News & 0.9812 & 0.8910 & 0.9009 & 0.0000  \\ 
\hline 
RTE & 0.9000 & 0.8118 & 0.3762 & 0.1980  \\ 
\hline 
\end{tabular} 
\vspace{0.1cm}
\caption{Hyper-parameters values for the decoder of UnifiedQA T5-Large.} 
\label{tab:hyper_parameters_UQA_encoder} 
\end{center} 
\end{table}

\vspace{-0.3cm}

\begin{table}[h!] 
\begin{center} 
\begin{tabular}{ || c || c | c | c | c ||}  
\hline 
\multirow{2}{*}{Dataset} & \multicolumn{4}{c||}{Loss Function} \\ 
\cline{2-5} 
  & MD-CCS & MD-Acc & MA & SMR \\ 
\hline 
\hline 
IMDB & 0.9891 & 0.9900 & 0.5940 & 0.4950 \\ 
\hline 
Amazon & 0.9891 & 0.8910 & 0.4950 & 0.6534  \\ 
\hline 
BoolQ & 0.9752 & 0.0000 & 0.0000 & 0.0990  \\ 
\hline 
AG News & 0.9891 & 0.0000 & 0.0000 & 0.8910  \\ 
\hline 
RTE & 0.9792 & 0.3366 & 0.3168 & 0.2772  \\ 
\hline 
\end{tabular} 
\vspace{0.1cm}
\caption{Hyper-parameters values  for the encoder of DeBERTa.} 
\label{tab:hyper_parameters_DeBERTa_encoder} 
\end{center} 
\end{table}

\vspace{-0.3cm}

\begin{table}[h!] 
\begin{center} 
\begin{tabular}{ || c || c | c | c | c ||}  
\hline 
\multirow{2}{*}{Dataset} & \multicolumn{4}{c||}{Loss Function} \\ 
\cline{2-5} 
  & MD-CCS & MD-Acc & MA & SMR \\ 
\hline 
\hline 
IMDB & 0.9713 & 0.0990 & 0.5742 & 0.4950 \\ 
\hline 
Amazon & 0.9911 & 0.9900 & 0.5940 & 0.5940  \\ 
\hline 
BoolQ & 0.9871 & 0.9900 & 0.9108 & 0.4950  \\ 
\hline 
AG News & 0.9871 & 0.8910 & 0.9702 & 0.9900  \\ 
\hline 
RTE & 0.9871 & 0.9900 & 0.9900 & 0.5940  \\ 
\hline 
\end{tabular}
\vspace{0.1cm}
\caption{Hyper-parameters values  for the decoder of GPT-Neo.} 
\label{tab:hyper_parameters_GPT-Neo_decoder} 
\end{center} 
\end{table}